\documentclass[11pt,a4paper]{article}
\usepackage[T1]{fontenc}
\usepackage[utf8]{inputenc}
\usepackage{authblk}
\usepackage[square,numbers]{natbib}
\usepackage[margin=1.0in]{geometry}

\usepackage{fancyhdr}

\usepackage{hyperref}
\usepackage{graphicx}

\fancypagestyle{plain}{%
  \fancyhf{}%
  \fancyfoot[C]{Copyright © 2021 for this paper by its authors. Use permitted under Creative Commons License Attribution 4.0 International (CC BY 4.0).\\\textit{Proceedings of the Conference on Applied Machine Learning for Information Security}, 2021}%
}

\usepackage{lipsum}

\title{Shell Language Processing: \\Unix command parsing for Machine Learning}
\author{Dmitrijs Trizna\\\texttt{dmitrijs.trizna@helsinki.fi}\\University of Helsinki\\Avast Software}
\date{} 

\begin{document}

\maketitle

\begin{abstract}
In this article, we present a Shell Language Preprocessing (SLP) library, which implements tokenization and encoding directed on parsing of Unix and Linux shell commands. We describe the rationale behind the need for a new approach with specific examples when conventional Natural Language Processing (NLP) pipelines fail. Furthermore, we evaluate our methodology on a security classification task against widely accepted information and communications technology (ICT) tokenization techniques and achieve significant improvement of an F1-score from 0.392 to 0.874.
\end{abstract}

\section{Introduction}

One of the most common interfaces that system engineers and administrators use to manage computers is the command line. Numerous interpreters called \emph{shells} allow applying the same operational logic across various sets of operating systems. Bourne Shell (abbreviated \emph{sh}), Debian Almquist shell (\emph{dash}), Z shell (\emph{zsh}), and Bourne Again Shell (\emph{bash}) are ubiquitous interfaces to operate Unix systems, known for speed, efficiency, ability to automate and integrate the diversity of tasks.

Contemporary enterprises rely on \texttt{auditd} telemetry from Unix servers, including \texttt{execve} syscall data describing executed shell commands. Generally, this data is analyzed using static, manually defined signatures for needs like intrusion detection. There is sparse evidence of the ability to use this information source efficiently by modern data analysis techniques, like machine learning or statistical inference. Leading ICT companies underline the relevance of the aforementioned problem and perform internal research to address this issue~\cite{fsecure}.

Complexity and malleability are key challenges in pre-processing shell commands. The structure has many intricate details, like aliases, different prefixes, and order and values of the text, which make commands condensed and fast to input, but time-consuming when reading and interpreting them.

In this paper, we present a \emph{Shell Language Processing (SLP)}\footnote{\url{https://github.com/dtrizna/slp}} library, showing how it performs feature extraction from raw shell commands and successfully use it as input for different machine learning models. The shell command-specific syntax and challenges are discussed in Section 2, inner working, and features of our pipeline are covered in Section 3. Sections 4 and 5 discuss a specific application case and provides performance evaluation of different encoding techniques.

\section{Shell specific challenges}

The syntax of shell commands depends on different Linux binary implementations and the way they handle input parameters. Standard techniques like the \textit{tokenize} method from the \emph{nltk} package would result in wrong tokens since shell language heavily deviates from natural language. To start with, spaces do not always separate different parts of the command. Some commands like \texttt{sed} take a raw regular expression as a parameter, space (and any other special character) will not necessarily mean the start of a next token:
\begin{verbatim}
  sed 's/^chr//;s/\..* / /' filename
\end{verbatim}

On the contrary, \texttt{java} takes flags without spaces at all, knowing where its parameter name ends:
\begin{verbatim}
  java -Xms256m -Xmx2048m -jar remoting.jar
\end{verbatim}

Such malleability in syntactic patterns possess a significant challenge for the successful parsing of shell command lines. We know two libraries that attempt to address such specifics for bash - \emph{bashlex}\footnote{\url{https://github.com/idank/bashlex}} and \emph{bashlint}\footnote{\url{https://github.com/skudriashev/bashlint}}, however none of them does its job perfectly. We utilize bashlex for our needs as a primary parsing source since it provides heuristics for tokenization of the most general patterns of shell commands. Still, there are multiple problems in the bashlex library’s original syntax analysis process, for example in the "command within a command" case (known by shell syntax \texttt{\$(cmd)} or \`{\texttt{cmd}}\`~) like in:

\begin{verbatim}
  export IP=$(dig +short example.com)
\end{verbatim}

Such syntax is not handled by bashlex, where an embedded command is treated as a single element. Therefore we implement additional syntactic logic wrapped around bashlex's object classes to handle this and other problematic cases.

\section{Encoding}

Subsequently, we decided to look forward to different ways of representing data numerically. To produce arrays out of textual data we implement (a) label, (b) one-hot, and (c) term frequency-inverse document frequency (TF-IDF) encodings. Label encoding is built on top of the scikit-learn \cite{sklearn_api}, whereas the one-hot and TF-IDF encodings are implemented natively.

Having the tokenization and encoding now as operational functionalities, we were ready to provide a user-friendly interface to utilize the underlying code. Therefore, we created dedicated Python classes that formed the core of our interface library. The tokenization interface is available via \textbf{ShellTokenizer()} class. Besides \emph{bashlex}, we utilized a \emph{Counter} object from the \textit{collections} package to store unique tokens and their appearance count. The Counter allows working conveniently with the data concerning various visualizations. Further encoding is available by \textbf{ShellEncoder()}, which class allows to generate different encoding methods and returns a \emph{scipy} sparse array. Substantially, all preprocessing pipeline can be achieved within four lines of code:

\begin{verbatim}
st = ShellTokenizer()
corpus, counter = st.tokenize(shell_commands)

se = ShellEncoder(corpus=corpus, 
                  token_counter=counter, 
                  top_tokens=100)
X_enc = se.tfidf()
\end{verbatim}

At this stage \texttt{X\_enc} can be supplied to \texttt{fit()} method of API-interface supported by machine learning (ML) models in libraries like scikit-learn~\cite{sklearn_api} or even AutoML realizations like TPOT~\cite{tpot}. As a result, every administrator, security analyst, or even non-technical manager working with a Unix-like system may parse shell data for ML-based analytics using our library. Only basic Python coding skills and no knowledge of conventional Natural Language Processing (NLP) pipelines are needed. 

\section{Experimental setup}

Assessment of tokenization and encoding quality is done on the security classification problem, where we train an ML model to distinguish malicious command samples from benign activity. Legitimate commands consist from \emph{nl2bash} dataset \cite{nl2bash}, which represents a shell commands collected from question-answering forums like \url{stackoverflow.com} or administratively focused cheat-sheets.

We perform the collection of malicious samples by ourselves, accumulating harmful examples across penetration testing and hacking resources that describe how to perform enumeration of Linux targets and acquire reverse shell connections from Unix hosts\footnote{For example \url{https://blog.g0tmi1k.com/2011/08/basic-linux-privilege-escalation/}}. All commands within the dataset are normalized. Domain names are replaced by \texttt{example.com} and all IP addresses with \texttt{1.1.1.1}, since in our evaluation we want to focus on the command interpretability. We advice to perform a similar normalization even in production applications, otherwise ML model will overfit training data.

For classification, we train a gradient boosting ensemble of decision trees, with the specific realization from XGBoost library \cite{Chen_2016}. During the analysis of tokenization we do not implement a validation, ergo use full dataset for training and compute metrics on the same data. Conversely, during the evaluation of encoding techniques, we implement a 10 fold cross-validation, so the resulting metrics are the mean across all validation passes.

\section{Evaluation}

First experiments were conducted to assess the quality of our tokenization. We have tokenized aforementioned dataset using our \texttt{ShellTokenizer}, with alternatives from NLTK \texttt{WordPunctTokenizer}, and \texttt{WhiteSpaceTokenizer} which are known to be used in ICT industry for log parsing. Then all the three corpora are encoded using the same term frequency-inverse document frequency (TF-IDF) realization. 

\begin{table}
\begin{center}
\begin{tabular}{ccccc}
Tokenizer & AUC & F1 & Precision & Recall \\
\hline \\
{\bf SLP (ours)}
& \bf{0.994} & \bf{0.874} & 0.980 & \bf{0.789} \\
{\bf WordPunct}
& 0.988 & 0.392 & \bf{1.0} & 0.244 \\
{\bf WhiteSpace}
& 0.942 & 0.164 & \bf{1.0} & 0.089 \\
\end{tabular}
\end{center}
\caption{Comparison of tokenization tecnhiques on security classification task: SLP, WordPunctTokenizer, WhiteSpaceTokenizer.}
\label{table:metrics}
\end{table}

Security classification results are available in Table \ref{table:metrics}. We see that conventional tokenization techniques fail to distinguish crucial syntactic patterns of commands, therefore both cases are biased towards the majority class of benign samples (with high precision and low recall). This situation is unacceptable for tasks like security classification - a model that fails to detect malicious samples at all (false-negatives) cannot be deployed in a production environment. On the contrary, our tokenizer achieves \emph{significantly} higher recall values, and, consequently, F1-score (0.874 versus 0.392 from the closest alternative).



We consider on the evaluation of different encoding techniques. Tests with our tokenization implementation are done against three encoding schemes - TF-IDF, one hot, and label encoding. Empirical evidence shows that none of the encoding techniques has a clear preference over the others, therefore, all three encoding types are implemented within the library since even basic logic like label encoding can yield the best results on specific data types and problem statements. We encourage analysts to experiment with various shell preprocessing techniques to understand which way benefits their pipelines.

\section{Conclusions}

In this article, we presented custom tokenization and encoding techniques focused on Unix shell commands. We describe the rationale of the dedicated tokenization approach, with specific shell command examples where conventional NLP techniques fail, and briefly cover the inner working of our library. To distinguish this technique from known existing pipelines we evaluate the security classification task, with a custom dataset collected from real-world samples across the web and an efficient ensemble model. According to acquired metrics, our model achieves a significant improvement of the F1-score.
\bibliography{citations}

\end{document}